\documentclass[]{style/style}
\usepackage{microtype}
\usepackage{graphicx}
\usepackage{subcaption}
\usepackage{csquotes}
\usepackage{afterpage}
\usepackage{xcolor}

\usepackage{amsmath}
\usepackage{amssymb}
\usepackage{mathtools}
\usepackage{amsthm}
\usepackage{xspace}
\usepackage{colortbl}
\usepackage{multirow}
\usepackage{multicol}
\usepackage{enumitem}
\usepackage{wrapfig}
\usepackage{nicefrac}
\usepackage[font=small,labelfont=bf,justification=justified,singlelinecheck=false]{caption}

\usepackage{booktabs}
\usepackage{multirow}
\usepackage{graphicx}
\usepackage{xcolor}
\usepackage{geometry}
\usepackage{array}
\usepackage{tabularx}
\newcolumntype{L}[1]{>{\raggedright\arraybackslash}p{#1}}
\newcolumntype{Y}{>{\raggedright\arraybackslash}X}

\usepackage{amsmath}
\title{Video = World + Event~Stream}

\author[]{Wan Team, Alibaba Group}

\affiliation[]{\small See \hyperref[app:contributions-and-acknowledgements]{Contributions and Acknowledgements} for the full author list.}

\abstract{
We present \textbf{Wan-Streamer v0.3}, which reframes our native-streaming interaction model under a single organizing view: \textbf{a video is a world plus an event stream}. The \emph{world} is the persistent context in which a video unfolds, including the environment, scene, subjects, ambient acoustic conditions, voice characteristics, and other relatively stable conditions. The \emph{event stream} is everything that changes over time within that world, including scene or environmental changes, subject behavior, speech, and other sounds. This yields a \textbf{general-purpose pretraining task} over large amounts of real video: given a world and incoming input, predict how the world moves, changes, and responds in real time. The resulting competence can be specialized to a broad family of real-time downstream tasks. We instantiate it on \textbf{real-time full-duplex audio-visual interaction}, where the event stream is the agent's speech together with free-form behavior. Functionally, the model's multimodal understanding process is vision-language-action-like: it maps multimodal user input to language-form speech and behavior actions. Wan-Streamer v0.3 preserves the v0.2 operating point: \textbf{640$\times$368} video at \textbf{25 FPS}, a \textbf{160 ms} streaming unit, approximately \textbf{200 ms} model-side response latency, and approximately \textbf{550 ms} total interaction latency under a 350 ms bidirectional network budget.
}

\metadata[Website]{\url{https://wan-streamer.com/}}
\begin{document}

\maketitle

\section{Introduction}
\label{section:intro}

Wan-Streamer v0.1~\citep{huang2026wanstreamerv01endtoendrealtime} established a native-streaming, end-to-end formulation for real-time full-duplex audio-visual interaction: user and agent text, audio, and video live on one causal timeline and are modeled by a single Transformer, so perception, response timing, speaking, visible listening, and synchronized video are learned as one behavior rather than assembled from cascaded modules. Wan-Streamer v0.2~\citep{huang2026wanstreamerv02higherresolution} kept that formulation fixed and raised the interactive stream from 192$\times$336 to 640$\times$368 at the same $\sim$200 ms model-side latency, moving the expensive latent generation into a Ulysses-style context-parallel performer~\citep{jacobs2023deepspeedulysses}.

Both versions treat one target application---an agent that talks and reacts---as the training objective. Wan-Streamer v0.3 steps back and asks what the underlying competence is. Our answer is a simple decomposition: \textbf{any video can be read as a world plus an event stream}. The \emph{world} is the persistent context in which a video unfolds: its environment and scene, visual style, subjects and their identities and appearances, ambient acoustic conditions, voice characteristics, and other relatively stable conditions. The \emph{event stream} is everything that changes over time within that context: scene or environmental changes, subject behavior, camera motion, speech, and other sounds. A recorded video is one realization of an event stream unrolling inside a given world; an interactive experience is the same object with the event stream produced online in response to input.

This decomposition defines a general-purpose pretraining task. We pretrain on large amounts of ordinary video to learn one thing: \textbf{given a world and an incoming input, how does the world stream forward---how does it move, change, and respond, unit by unit, in real time.} Because essentially all natural video is an instance of a world with an event stream, this objective is broad and scalable, and the competence it produces is world knowledge about how scenes plausibly evolve. That competence can then be specialized by varying the world context and the input that drives the stream, while retaining the same streaming response logic.

\begin{figure}[t]
    \centering
    \includegraphics[width=\linewidth]{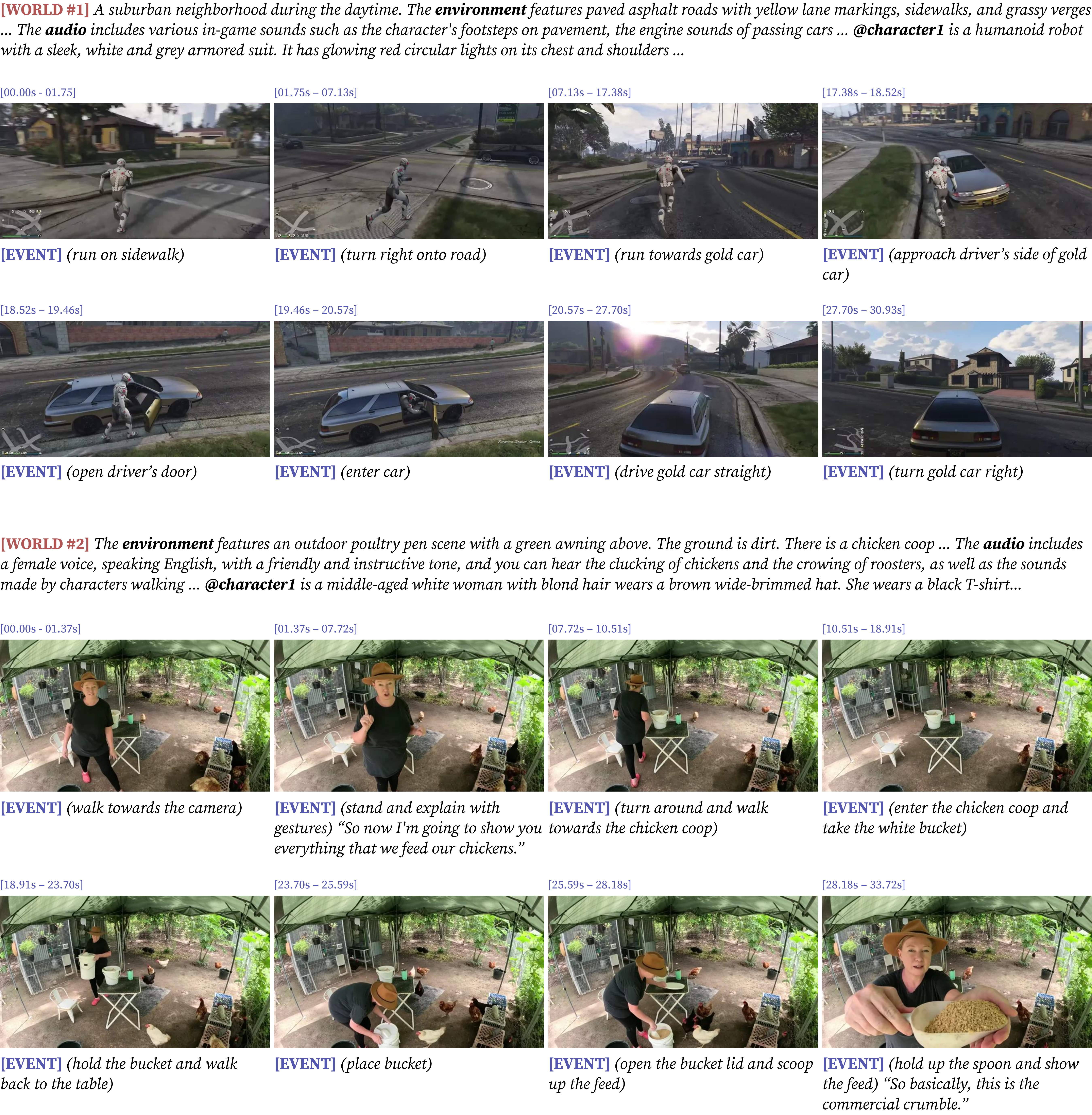}
    \caption{Two examples of the general-purpose pretraining data: time-aligned events paired with a summary of the persistent world context, including the visual setting, acoustic conditions, and characters. Parentheses denote character behavior, while quotation marks enclose spoken words. For clarity, both examples contain only one character, so actor names are omitted from the event labels. Training supports multiple characters by associating each event with its corresponding character, or with an explicit no-character marker for character-independent events.}
    \label{fig:pretraining-event-stream}
\end{figure}

We instantiate the pretrained world-event model on the same downstream task as v0.1/v0.2---real-time full-duplex audio-visual interaction---but widen what the agent can express. Here the \emph{world} configures the scene, character, ambient sound, and voice; the \emph{event stream} is the agent's response, driven by the user's streaming text, audio, and video. The key change from v0.2 is that the agent's event stream now carries \textbf{free-form behavior} alongside speech. We adopt a role-play chat format in which the model's language/state stream interleaves spoken words with parenthesized behavior directives, e.g. \enquote{\emph{(frowning slightly)} What did you say?} or \enquote{\emph{(picks up the mug and glances toward the window)} Sure, one moment.} The parenthesized part can be any behavior describable in language, so the agent is not limited to a small fixed action vocabulary (such as the $\mathtt{w}/\mathtt{a}/\mathtt{s}/\mathtt{d}$ moves used in roaming) but can perform open-vocabulary actions grounded in the world. This interleaved stream conditions audio-video generation exactly where the language stream did before, so no new latency-critical path is introduced.

\begin{figure}[t]
    \centering
    \includegraphics[width=\linewidth]{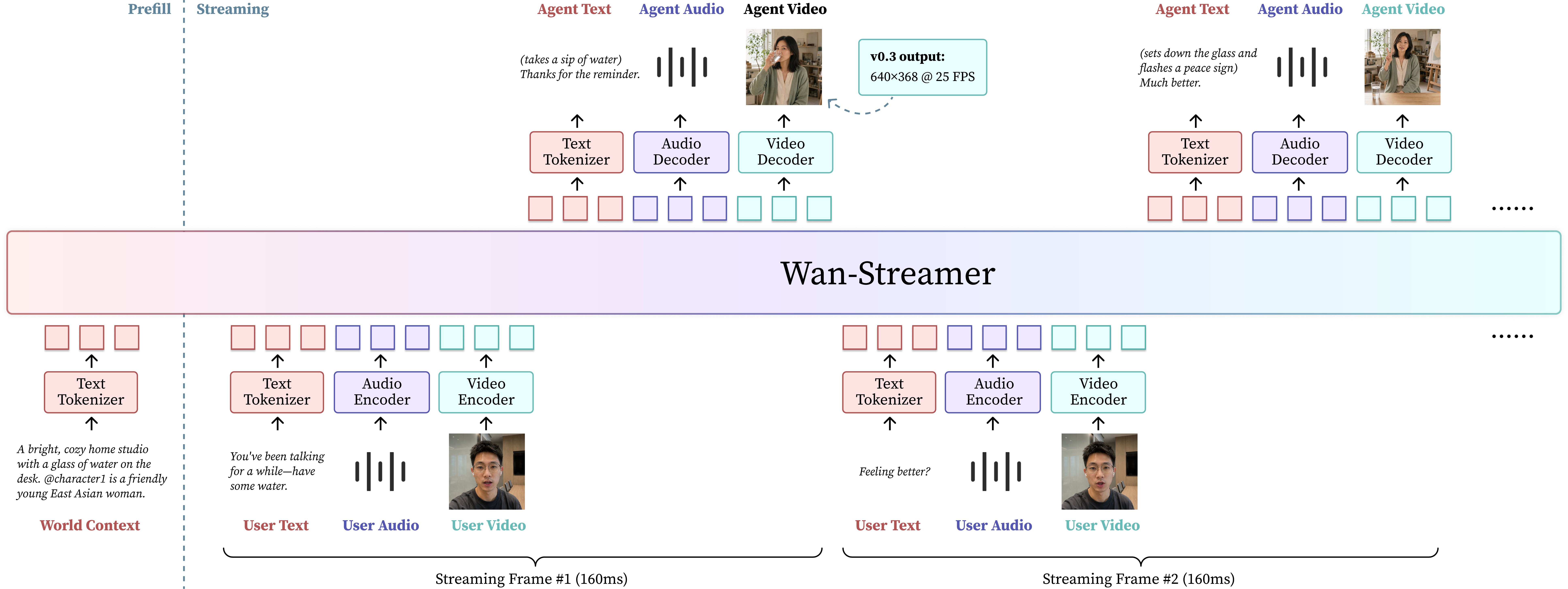}
    \caption{The world context is tokenized and prefilled once before streaming. Language, audio, and video inputs and outputs then share a causal timeline coordinated by block-causal attention. The model maps this world and streaming multimodal user input to language-form speech and behavior actions. These predicted tokens condition synchronized audio-video generation. In the agent text stream, phrases inside parentheses denote free-form behavior, while text outside parentheses is spoken by the agent.}
    \label{fig:framework}
\end{figure}

Because the modeling contract and the deployment topology are unchanged, v0.3 preserves the v0.2 operating point: 640$\times$368 video at 25 FPS, a 160 ms streaming unit, approximately 200 ms model-side response latency, and approximately 550 ms total interaction latency with a 350 ms bidirectional network budget. Figure~\ref{fig:pretraining-event-stream} shows the general pretraining representation before Figure~\ref{fig:framework} shows its interaction instantiation. The rest of the paper formalizes the decomposition and objective, describes the downstream behavior interface, compares versions, and reports experiments. The serving topology is inherited unchanged from v0.2, so we do not restate it here.

Our contributions are:
\begin{itemize}[leftmargin=1.3em]
    \item We reframe native-streaming generation as a \textbf{world + event stream} decomposition that separates persistent world context from time-varying events, yielding a general-purpose pretraining objective that transfers across real-time tasks.
    \item We instantiate this model on real-time full-duplex audio-visual interaction, where the model's multimodal understanding is VLA-like, mapping streaming multimodal input to speech and \textbf{open-vocabulary behavior actions expressed in natural language}. These language-form outputs condition synchronized audio-video generation.
    \item We show that this expansion keeps the v0.1/v0.2 modeling contract and the v0.2 serving topology intact, preserving 640$\times$368 at 25 FPS with approximately 200 ms model-side and approximately 550 ms total interaction latency.
\end{itemize}

\clearpage

\section{World-Event Decomposition}
\label{sec:world-event}

\subsection{Formalization}
We read a video as a persistent \emph{world} together with a time-varying \emph{event stream}. The world provides the context that remains relatively stable over the timescale of a clip or interaction---its visual setting and style, environment and layout, subjects and their identities and appearances, ambient acoustic conditions, and voice characteristics. The event stream contains the changes that unfold within that context, including subject behavior, camera motion, environmental change, speech, and other sound events.

For learning, we represent the world as a flexible structured record
\begin{equation}
    W = \{w_j\}_{j \in \mathcal{S}(W)},
\end{equation}
where the schema $\mathcal{S}(W)$ is determined by the clip or task rather than fixed globally. In practice, fields commonly group visual context (medium or style, environment, layout), acoustic context (ambient sound and voice characteristics), and a set of character records (identity, appearance, persona, visibility), but fields may be absent or extended.

An event is a time-localized record
\begin{equation}
    e_k = \big(\tau_k,\, c_k,\, d_k\big), \qquad c_k \in \mathcal{C}(W) \cup \{\varnothing\},
\end{equation}
where $\tau_k$ is its active interval, $d_k$ is a free-form description of the change, and $c_k$ associates the event with a character in $W$ or explicitly marks it as character-independent. The description may cover behavior, camera motion, environmental change, speech, or physical sound; overlapping changes can be represented by multiple records. A video is then a world with its event stream unrolled, and its observable audio-visual realization is produced by the causal decoders of v0.1 from the generated latents. This is not a new architecture: it is a reading of the same interleaved causal sequence in which durable conditioning is named $W$ and time-localized change is named $e_k$.

\subsection{Streaming world-event factorization}
The v0.1 formulation predicts the next response unit from the full causal history. Writing that response as the event unit $e_k$ and letting $x_{\le k}$ denote whatever input drives the stream (for pretraining, the observed past; for interaction, the user's streaming observations), the model factorizes as
\begin{equation}
\label{eq:world-event-factorization}
    p_\theta\big(e_{1:K}\mid W, x_{1:K}\big) = \prod_{k=1}^{K} p_\theta\big(e_k \mid W,\, x_{\le k},\, e_{<k}\big),
\end{equation}
where $K$ is the number of streaming units. The v0.1 autoregressive factorization is the special case in which $W$ is folded into the history, $x$ is the user observation, and $e_k$ is the agent's text/audio/video response. As before, discrete parts of $e_k$ (language, and now behavior directives) are optimized with next-token prediction, while the continuous audio and video latents are generated with conditional flow matching under the same clean context, so speech, motion, appearance, and scene evolution are denoised as one coupled response and committed back into history for the next unit. Naming $W$ explicitly makes the persistent world context a first-class conditioning object: at inference it is provided once and the event stream is produced online.

\subsection{Pretraining and downstream transfer}
Equation~\eqref{eq:world-event-factorization} defines a general-purpose pretraining task over large-scale, diverse video corpora rather than one interaction domain. Each clip supplies an inferred world and time-aligned events, and the model predicts the next event and its audio-visual realization from causal history. This competence can in principle support many downstream interfaces, each defined by its world context and stream-driving input: world-model control and real-time first-person interaction use action or navigation input over a scene world~\citep{worldplay2025,matrixgame2026,cheng2026360explorer,realwonder2026}; embodied manipulation uses commands over a workspace world~\citep{ao2024bodyofher,lingbotva2026,flowact2026}; and our instantiation uses streaming user text, audio, and video over a scene-and-character world. This paper studies only the real-time full-duplex audio-visual interaction specialization and leaves systematic adaptation and evaluation of the others to future work.

Figure~\ref{fig:pretraining-event-stream} makes this supervision concrete: each example summarizes its persistent world context once using the information relevant to that clip, while natural-language events are aligned to the intervals in which they occur. Learning how these events unfold in context teaches the pretrained model world-event dynamics that can be reused when events are produced online at inference.

\FloatBarrier

\section{Real-time Interaction}
\label{sec:downstream}

\subsection{Instantiating the world and the event stream}
For real-time full-duplex audio-visual interaction, the persistent world context is instantiated with the settings needed by this task:
\begin{itemize}[leftmargin=1.3em]
    \item \textbf{Scene setting}: the environment and layout the agent inhabits.
    \item \textbf{Character setting}: the agent's identity, appearance, and persona.
    \item \textbf{Ambient-sound setting}: the background acoustic field of the scene.
    \item \textbf{Voice timbre}: the agent's speaking voice.
\end{itemize}
The user's streaming text, audio, and video are the input $x_{\le k}$ that drives the event stream. In v0.3, the agent response interleaves speech and free-form behavior records, both rendered into synchronized audio and video by the causal decoders.

This inference-time behavior stream is the controllable counterpart of the time-aligned behavior language used during pretraining in Figure~\ref{fig:pretraining-event-stream}: instead of being recovered from recorded video, the behavior event is produced or supplied online and the model streams the corresponding world evolution.

\subsection{Free-form behavior in a role-play format}
v0.1 and v0.2 focus on an agent that speaks and shows visible listening behavior---idle presence, gaze, nods, micro-expressions, lip-synchronized speech. v0.3 widens the behavior channel to \textbf{open-vocabulary actions described in natural language}. We adopt a role-play chat format in which the agent's language output interleaves spoken words with parenthesized behavior directives, for example:
\begin{quote}
\emph{(reaches into the grass and picks up a green leaf)} Look what I found.\\
\emph{(covers mouth in surprise)} I did not expect that.
\end{quote}
The parenthesized directive can be any behavior expressible in language, so the agent is not confined to a small fixed action set (such as the $\mathtt{w}/\mathtt{a}/\mathtt{s}/\mathtt{d}$ navigation moves used in roaming) but can perform situated, open-vocabulary actions that are grounded in the configured world---reaching for a nearby object, turning toward a sound, changing posture, or reacting expressively. Behavior directives and spoken words share one token stream, so their timing, ordering, and interleaving with speech are learned jointly rather than scheduled by an external controller.

\subsection{Multimodal understanding and language-form actions}
Functionally, the model's multimodal understanding process is vision-language-action-like: conditioned on the world, it continually maps the user's text, audio, and video to language-form speech and behavior actions. These actions form an interleaved speech-and-behavior token stream that conditions joint audio-video latent generation. Behavior directives are short language tokens, so they add negligible cost to the token-causal path. The resulting behavior is realized visually (posture, gesture, gaze, interaction with the scene) and acoustically (prosody, non-verbal sound), keeping speech, behavior, and appearance synchronized before decoding rather than aligned afterwards.

\FloatBarrier

\section{Version Comparison}
\label{sec:upgrade}

Table~\ref{tab:v03-upgrade} summarizes the changes across versions. The end-to-end streaming formulation, the resolution and frame rate, the latency budget, and the serving topology are inherited from v0.2; v0.3 changes the \emph{conceptual framing} (world + event stream), the \emph{pretraining objective} (predict how a world streams forward), and the \emph{agent's expressive range} (speech plus free-form behavior).

\begin{table}[!ht]
    \caption{Summary of the Wan-Streamer versions. Emphasized cells indicate what v0.3 changes; unemphasized cells indicate what it preserves from v0.2.}
    \label{tab:v03-upgrade}
    \centering
    \footnotesize
    \setlength{\tabcolsep}{5pt}
    \renewcommand{\arraystretch}{1.16}
    \begin{tabularx}{\textwidth}{@{}L{.17\textwidth}L{.20\textwidth}L{.22\textwidth}Y@{}}
        \toprule
        Aspect & v0.1 & v0.2 & v0.3 \\
        \midrule
        Framing & End-to-end streaming A/V interaction & Same, higher resolution & \textbf{Video = world + event stream} \\
        Core objective & Interaction data & Interaction data & \textbf{General-purpose world-event pretraining, then specialization} \\
        Output resolution & 192$\times$336 & 640$\times$368 & 640$\times$368 \\
        Frame rate & 25 FPS & 25 FPS & 25 FPS \\
        Model-side latency & $\sim$200 ms & $\sim$200 ms & $\sim$200 ms \\
        Total latency & $\sim$550 ms (350 ms network) & $\sim$550 ms (350 ms network) & $\sim$550 ms (350 ms network) \\
        Serving & Thinker + single-GPU performer & Thinker + Ulysses context-parallel performer & Same as v0.2 \\
        Agent event stream & Speech + visible listening behavior & Speech + listening, higher fidelity & \textbf{Speech + open-vocabulary free-form behavior} \\
        Behavior format & Implicit listening/idle motion & Implicit listening/idle motion & \textbf{Role-play with parenthesized behavior directives} \\
        \bottomrule
    \end{tabularx}
\end{table}

\FloatBarrier

\section{Experiments}
\label{sec:experiments}

\textbf{Latency and runtime protocol.}
We use the same response boundary as v0.1/v0.2. Model-side signal-to-signal latency starts when a 160 ms user streaming unit is available to the thinker and ends when the corresponding audio-video response unit has been decoded for emission. Because the serving topology is inherited unchanged from v0.2, v0.3 keeps approximately 200 ms model-side latency while producing 640$\times$368 video at 25 FPS, and approximately 550 ms total interaction latency with a 350 ms bidirectional network budget. We keep the network term as an external deployment assumption so the comparison isolates the model-side path, and we report the v0.3 runtime at the same boundary used in prior versions~\citep{doubao_realtime_voice2025,openai2024gpt4o,hume2025evi3,streamavatar2025,hallolive2026}.

\textbf{Qualitative behavior observations.}
In generated 640$\times$368 conversations, Wan-Streamer v0.3 speaks while performing open-vocabulary behavior in real time. Behavior directives in the language stream are realized as coherent facial expressions, posture changes, responses to sounds, and interactions with nearby objects. These actions remain synchronized with speech, grounded in the configured scene, and consistent with the character's identity and appearance. Between explicit behavior events, the agent maintains stable idle and listening behavior. Wan-Streamer v0.3 therefore expands the agent's expressive range beyond speech and implicit listening while retaining the same low-latency streaming setting as v0.2.

\section{Conclusion}

Wan-Streamer v0.3 reframes native-streaming generation as a \textbf{world + event stream} decomposition: persistent world context plus everything that changes over time. This supports general-purpose video pretraining and transfer to roaming, audio-visual interaction, and embodied manipulation. We instantiate this framework through downstream post-training for real-time full-duplex audio-visual interaction. In this instantiation, VLA-like multimodal understanding maps streaming user input to language-form speech and open-vocabulary behavior actions, which condition synchronized audio-video generation. Wan-Streamer v0.3 retains v0.2's modeling contract and serving topology, preserving 640$\times$368 video at 25 FPS, with approximately 200 ms model-side and 550 ms total interaction latency.

\clearpage
\bibliographystyle{unsrtnat}
\bibliography{paper}

@misc{huang2026wanstreamerv02higherresolution,
  title         = {{Wan-Streamer v0.2: Higher Resolution, Same Latency}},
  author        = {Lianghua Huang and Zhi-Fan Wu and Yupeng Shi and Wei Wang and
                   Mengyang Feng and Junjie He and Chen-Wei Xie and Yu Liu and
                   Jingren Zhou and Ang Wang and Bang Zhang and Baole Ai and
                   Chen Liang and Cheng Yu and Chongyang Zhong and Jinwei Qi and
                   Kai Zhu and Pandeng Li and Peng Zhang and Wenyuan Zhang and
                   Xinhua Cheng and Yitong Huang and Yun Zheng and Yuxiang Bao and
                   Yuzheng Wang and Zoubin Bi},
  year          = {2026},
  month         = jul,
  eprint        = {2607.04443},
  archivePrefix = {arXiv},
  primaryClass  = {cs.CV},
  url           = {https://arxiv.org/abs/2607.04443},
  doi           = {10.48550/arXiv.2607.04443},
  note          = {Submitted on 5 Jul 2026}
}

@misc{huang2026wanstreamerv01endtoendrealtime,
  title         = {{Wan-Streamer v0.1: End-to-end Real-time Interactive Foundation Models}},
  author        = {Lianghua Huang and Zhi-Fan Wu and Wei Wang and Yupeng Shi and
                   Mengyang Feng and Junjie He and Chen-Wei Xie and Yu Liu and
                   Jingren Zhou and Ang Wang and Bang Zhang and Baole Ai and
                   Chen Liang and Cheng Yu and Chongyang Zhong and Jinwei Qi and
                   Kai Zhu and Pandeng Li and Peng Zhang and Wenyuan Zhang and
                   Xinhua Cheng and Yitong Huang and Yun Zheng and Zoubin Bi},
  year          = {2026},
  month         = jun,
  eprint        = {2606.25041},
  archivePrefix = {arXiv},
  primaryClass  = {cs.CV},
  url           = {https://arxiv.org/abs/2606.25041},
  doi           = {10.48550/arXiv.2606.25041},
  note          = {Submitted on 23 Jun 2026}
}

@misc{doubao_realtime_voice2025,
  title        = {Doubao Realtime Voice Model},
  author       = {{ByteDance Seed Team}},
  howpublished = {\url{https://seed.bytedance.com/en/realtime_voice}},
  note         = {Model page, January 20, 2025},
  year         = {2025}
}

@misc{openai2024gpt4o,
  title        = {Hello GPT-4o},
  author       = {{OpenAI}},
  howpublished = {\url{https://openai.com/index/hello-gpt-4o/}},
  note         = {Blog post, May 13, 2024},
  year         = {2024}
}

@misc{hume2025evi3,
  title        = {Introducing EVI 3: The World's Most Realistic and Instructible Speech-Language Model},
  author       = {{Hume AI}},
  howpublished = {\url{https://www.hume.ai/blog/introducing-evi-3}},
  note         = {Blog post, 2025},
  year         = {2025}
}

@article{ao2024bodyofher,
  title   = {Body of Her: A Preliminary Study on End-to-End Humanoid Agent},
  author  = {Ao, Tenglong},
  journal = {arXiv preprint arXiv:2408.02879},
  year    = {2024}
}

@article{streamavatar2025,
  title   = {StreamAvatar: Streaming Diffusion Models for Real-Time Interactive Human Avatars},
  author  = {Sun, Zhiyao and Peng, Ziqiao and Ma, Yifeng and Chen, Yi and Zhou, Zhengguang and Zhou, Zixiang and Zhang, Guozhen and Zhang, Youliang and Zhou, Yuan and Lu, Qinglin and Liu, Yong-Jin},
  journal = {arXiv preprint arXiv:2512.22065},
  year    = {2025}
}

@article{flowact2026,
  title   = {FlowAct-R1: Towards Interactive Humanoid Video Generation},
  author  = {Wang, Lizhen and Zhu, Yongming and Ge, Zhipeng and Zheng, Youwei and Zhang, Longhao and Hu, Tianshu and Qin, Shiyang and others},
  journal = {arXiv preprint arXiv:2601.10103},
  year    = {2026}
}

@article{hallolive2026,
  title   = {Hallo-Live: Real-Time Streaming Joint Audio-Video Avatar Generation with Asynchronous Dual-Stream and Human-Centric Preference Distillation},
  author  = {Li, Chunyu and Li, Jiaye and Mei, Ruiqiao and Xia, Haoyuan and Zhu, Hao and Wang, Jingdong and Zhu, Siyu},
  journal = {arXiv preprint arXiv:2604.23632},
  year    = {2026}
}

@article{worldplay2025,
  title   = {WorldPlay: Towards Long-Term Geometric Consistency for Real-Time Interactive World Modeling},
  author  = {Sun, Wenqiang and Zhang, Haiyu and Wang, Haoyuan and Wu, Junta and Wang, Zehan and Wang, Zhenwei and Wang, Yunhong and Zhang, Jun and Wang, Tengfei and Guo, Chunchao},
  journal = {arXiv preprint arXiv:2512.14614},
  year    = {2025}
}

@article{matrixgame2026,
  title   = {Matrix-Game 3.0: Real-Time and Streaming Interactive World Model with Long-Horizon Memory},
  author  = {Wang, Zile and Liu, Zexiang and Li, Jiaxing and Huang, Kaichen and Xu, Baixin and Kang, Fei and An, Mengyin and others},
  journal = {arXiv preprint arXiv:2604.08995},
  year    = {2026}
}

@article{realwonder2026,
  title   = {RealWonder: Real-Time Physical Action-Conditioned Video Generation},
  author  = {Liu, Wei and Chen, Ziyu and Li, Zizhang and Wang, Yue and Yu, Hong-Xing and Wu, Jiajun},
  journal = {arXiv preprint arXiv:2603.05449},
  year    = {2026}
}

@article{lingbotva2026,
  title   = {Causal World Modeling for Robot Control},
  author  = {Li, Lin and Zhang, Qihang and Luo, Yiming and Yang, Shuai and Wang, Ruilin and Han, Fei and Yu, Mingrui and Gao, Zelin and Xue, Nan and Zhu, Xing and Shen, Yujun and Xu, Yinghao},
  journal = {arXiv preprint arXiv:2601.21998},
  year    = {2026}
}

@article{jacobs2023deepspeedulysses,
  title={DeepSpeed Ulysses: System Optimizations for Enabling Training of Extreme Long Sequence Transformer Models},
  author={Jacobs, Sam Ade and Tanaka, Masahiro and Zhang, Chengming and Zhang, Minjia and Song, Shuaiwen Leon and Rajbhandari, Samyam and He, Yuxiong},
  journal={arXiv preprint arXiv:2309.14509},
  year={2023}
}

@inproceedings{cheng2026360explorer,
  title={360Explorer: Exploring 4D Controllable World in Panoramic Videos},
  author={Cheng, Xinhua and Zhou, Haiyang and Yu, Wangbo and Jia, Tanghui and Lin, Bin and Ge, Yunyang and Li, Weiqi and Yuan, Li},
  booktitle={Proceedings of the AAAI Conference on Artificial Intelligence},
  volume={40},
  number={5},
  pages={3300--3308},
  year={2026}
}

\clearpage

\appendix
\section*{Appendix}

\section{Contributions and Acknowledgements}
\label{app:contributions-and-acknowledgements}

\subsection{Core Contributors}
Lianghua Huang\footnote{Corresponding author: \href{mailto:lianghua.huang.cs@gmail.com}{\nolinkurl{lianghua.huang.cs@gmail.com}}.}, Zhi-Fan Wu, Yupeng Shi, Wei Wang, Mengyang Feng, Cheng Yu, Chen Liang, Junjie He, Chen-Wei Xie, Yu Liu, and Jingren Zhou.

\subsection{Contributors}
Contributors are listed alphabetically by first name: Ang Wang, Bang Zhang, Baole Ai, Chongyang Zhong, Jinwei Qi, Kai Zhu, Pandeng Li, Peng Zhang, Wenyuan Zhang, Xinhua Cheng, Yitong Huang, Yun Zheng, Yuxiang Bao, Yuzheng Wang, Zhiwei Lin, and Zoubin Bi.

\end{document}